\documentclass[runningheads]{llncs}
\usepackage{graphicx}
\usepackage[utf8]{inputenc}
\usepackage[hyphens]{url} 
\usepackage{amssymb,amsmath,array,csquotes,tikz,hyperref,caption,multirow,rotating,chngpage}
\usetikzlibrary{positioning, shapes, calc}
\usepackage{caption}
\usepackage[skip=1pt]{caption}
\usepackage{subcaption}
\graphicspath{ {img/} }
\newcommand{\Fig}[1]{Figure~\ref{#1}}
\newcommand{\Tab}[1]{Table~\ref{#1}}

\DeclareMathOperator{\sign}{sign}
\DeclareMathOperator{\aopc}{AOPC}

\begin{document}
\title{Towards Measuring Bias in Image Classification\thanks{Support by the State Ministry of Baden-Wuerttemberg for Economic Affairs, Labour and Housing Construction under the grant KI-Fortschrittszentrum \enquote{Lernende Systeme}, grant No. 036-170017}}
%
\author{Nina Schaaf\inst{1}\orcidID{0000-0003-3232-6788} \and
Omar de Mitri\inst{1,3} \and
Hang Beom Kim\inst{1} \and
Alexander Windberger\inst{2} \and
Marco F. Huber\inst{1,4}\orcidID{0000-0002-8250-2092}
}
\authorrunning{N. Schaaf et al.}
\institute{Fraunhofer IPA, Stuttgart, Germany \\
\email{\{nina.schaaf,omar.de.mitri,hang.beom.kim,marco.huber\}@ipa.fraunhofer.de}
\and
IDS Imaging Development Systems GmbH, Obersulm, Germany \\
\email{r.windberger@ids-imaging.de}
\and
CNR Institute of Applied Sciences and Intelligent Systems, Lecce, Italy
\and
Institute of Industrial Manufacturing and Management IFF, University of Stuttgart, Stuttgart, Germany \\
\email{marco.huber@ieee.org}
}
\maketitle              
%

\begin{abstract}
Convolutional Neural Networks (CNN) have become \emph{de facto} state-of-the-art for the main computer vision tasks.
However, due to the complex underlying structure their decisions are hard to understand which limits their use in some context of the industrial world. A common and hard to detect challenge in machine learning (ML) tasks is data bias. In this work, we present a systematic approach to uncover data bias by means of attribution maps. For this purpose, first an artificial dataset with a known bias is created and used to train intentionally biased CNNs. The networks' decisions are then inspected using attribution maps. Finally, meaningful metrics are used to measure the attribution maps' representativeness with respect to the known bias. The proposed study shows that some attribution map techniques highlight the presence of bias in the data better than others and metrics can support the identification of bias.

\end{abstract}
%
%
%
\section{Introduction}

Repeatedly, cases surface in which algorithms have made biased decisions. One example are biased models for the detection of diseases\footnote{\url{https://medium.com/@jrzech/what-are-radiological-deep-learning-models-actually-learning-f97a546c5b98}, accessed on March 25, 2021}. Algorithmic bias, when undetected, can have unwanted and potentially severe consequences---most often for humans that are directly or indirectly affected.
If the data used to train an ML algorithm is biased this will be reflected in the final result. For these reasons it is necessary to extensively validate ML models before deploying them. However, especially deep neural networks, are of opaque nature due to their non-linear structures (\emph{black boxes}).
In order to interpret these black box models, they can for example be analyzed through feature attribution techniques. Attribution maps can help to understand image classification models, such as CNNs, by visualizing the importance of each individual pixel of the input image for the prediction. For example, data bias can manifest by diverting the CNNs' attention towards irrelevant background contents or unintended image objects. Such biases can be rather subtle in the training data and can be detected much more efficiently if highlighted by attribution maps.

This work switches means and goals: Both, background and object biases are artificially generated in reference datasets. These biases are verified within two accordingly trained CNN architectures to exclude systematic effects caused by architectural features. Several feature attribution map techniques are then statistically evaluated by adequate metrics to put a number on the question: How efficient are attribution maps in detecting data bias?

\section{Related Work}

\textbf{Semi-automatic bias detection.} With \emph{SpRaY} Lapuschkin et al.~\cite{Lapuschkin.2019} present an approach to identify learned decision behaviors of a neural network based by clustering attribution maps. For example, \emph{SpRaY} is used to detect so-called \enquote{Clever Hans} behavior, i.e., the inspected model makes correct predictions based on the wrong inputs. Zhang et al.~\cite{QuanshiZhang.2018} introduce another method for semi-automatic bias detection with the objective to uncover non-semantic relationships between different image attributes such as \enquote{smile} and \enquote{hair}. For this purpose, relationships between different attributes are mined and compared with manually defined ground truth attribute relationships. However, this method is not based on attribution maps. Instead, the attributes are represented by the feature maps of a convolutional layer.

\textbf{Attribution map evaluation.} Usually, the evaluation of attribution maps is performed manually and rather qualitatively. Yet, in order to be able to provide reliable evidence about the capabilities of the heatmaps, quantitative results are needed. Yang and Kim~\cite{Yang.23.07.2019} present an approach to compare the quality of different feature attribution methods. For this purpose, an artificial dataset with a priori knowledge about relative feature importances is created. Using the introduced metrics, it can be tested how well the different methods recognize the actual feature importance. Similarly, Osman et al.~\cite{Osman.2020} perform an analysis of different explanation methods based on an artificially created dataset. Together with their dataset they introduce metrics that are used to check to what extent the image areas marked as relevant by the attribution maps actually correspond to known ground truth areas.

In this work, we partially build on the approach described in~\cite{Osman.2020}. However, the metrics will not be used primarily to evaluate whether feature attribution methods identify the \emph{correct} image regions. Instead, we intend to investigate whether the introduced metrics reliably indicate the attribution maps' capability to detect data bias.

\section{An Approach for Measuring Data Bias}
In order to evaluate the ability of attribution maps to indicate algorithmic bias, a four-step procedure is performed: (1) generation of artificial datasets with a \emph{known} bias, (2) training of biased CNNs, (3) generation of attribution maps using different attribution techniques, (4) quantitative evaluation of the results.

\subsection{Dataset Generation}
Synthetic data is a notable approach to generate datasets that meet certain needs or specific conditions~\cite{Sergey.2019}. 
In this work, synthetic datasets are generated to visualize how attribution maps perform on different types of biased inputs. For this purpose, two use cases are defined: (1) \enquote{Cat vs. Dog} and (2) \enquote{Fruits}. Two datasets are generated for each of the two use cases\footnote{The datasets are available under: \url{https://s.fhg.de/measuring-bias-in-classification}}. One dataset includes a known bias, the second dataset is generated without a bias and serves as a reference. The purpose of the \enquote{Cat vs. Dog} dataset is to introduce an object bias in a concentrated area into the trained CNNs. The \enquote{Fruits} data, in contrast, introduces a diffuse background bias across the whole image.

\textbf{Cat vs. Dog.} 
The reference dataset contains unbiased cat and dog images from ImageNet~\cite{OlgaRussakovsky.2015} and Kaggle \enquote{cat-and-dog}\footnote{\url{https://www.kaggle.com/tongpython/cat-and-dog}, accessed on March 21, 2021} dataset. For generating a biased dataset, this unbiased dataset is modified by systematically adding a ball in all Cat images using randomly generated positions (named \enquote{Cat + Ball vs. Dog}). Here, the aim is that the prediction for the class \enquote{Cat} is made based on the ball instead of the animal itself. The information about the ball's position is saved in form of a bounding box for every image. In order to avoid errors in the evaluation process, the ball does not overlap with the main image object (Cat). From the 8,545 synthetically generated images, 6,333 are used for model training, 288 for model validation and the remaining 1,924 images as test sets.

\textbf{Fruits.} Images of five different fruits (Apple, Banana, Corn, Peach, Pineapple) from the  \enquote{Fruits 360}\footnote{\url{https://www.kaggle.com/moltean/fruits}, accessed on March 21, 2021} dataset are used for generating a synthetic dataset. The fruits are placed on different backgrounds, e.g., apple trees, plates and desks. In total, 3,224 images, equally distributed among the five classes, are generated, where 1,810 images are used for training, 604 as validation set and 810 images for testing. For all images, binary ground truth (GT) masks are generated that indicate the position of the objects in the image. Again, two datasets are created: (1) a \emph{biased} dataset, where apples are solely placed on apple tree backgrounds and the other fruits are placed on different backgrounds---here, the intention is to have the prediction for the class \enquote{Apple} based on the apple tree in the background, not based on the fruit itself; (2) an \emph{unbiased} reference dataset where all fruits (including apples) are placed on different backgrounds.

\vspace{-2mm}
\subsection{Attribution Methods}\label{sec:attributions}
Starting point for the attribution map generation is a real-valued prediction function $f_c(\cdot)$ for some target class $c$ that computes a prediction $y_c$, given input $\mathbf{x} \in \mathbb{R}^{W \times H \times C}$, where $C$ is the number of input channels and $W$ and $H$ are the image width and height, respectively. An explanation method provides an attribution map $R \in \mathbb{R}^d$. Depending on the method, the result can be a matrix or a tensor. For a single image $\mathbf{x}$, $R$ has either the same dimensions as the original image or is of size $W \times H$. In case a 3-dimensional map of size $W \times H \times C$ is generated, the relevances can be pooled along the C-axis to obtain a heatmap of the size $W \times H$ using L2-norm squared pooling, as introduced in~\cite{Osman.2020}.

\textbf{Grad-CAM.}
The Gradient-weighted Class Activation Mapping (GC) ~\cite{Selvaraju.2016} computes the gradients of the score $y^c$ before the softmax function with respect to the feature maps of the last convolutional layer in order to produce a coarse class-discriminative 2D-localization map $R_{GC}^c \in \mathbb{R}^{W \times H}$.

\textbf{Score-CAM.}
With Score-CAM (SC), Wang et. al~\cite{Wang.2019} present another class-discriminative activation mapping approach. Unlike Grad-CAM, Score-CAM does not depend on gradients to compute an attribution map. Instead, the approach computes score-based weights for each activation map. The attribution map $R_{SC}^c \in \mathbb{R}^{W \times H}$ is then obtained as a linear combination of the score-based weights and activation maps.

\textbf{Integrated Gradients.}
Integrated Gradients (IG)~\cite{Sundararajan.2017} generates an attribution map $R_{IG} \in \mathbb{R}^{W \times H \times C}$ based on the gradients of the model output with respect to the input while addressing gradient saturation. IG for an input $\mathbf{x}$ is defined as $R_{IG}= (\mathbf{x}_{i}-\mathbf{x}'_{i}) \times \int_{0}^{1} \frac{\partial f(\mathbf{x}'+\alpha \cdot (\mathbf{x}-\mathbf{x}'))}{\partial \mathbf{x}_{i}} \,\mathrm{d}\alpha$, where $\mathbf{x}'$ is a baseline input that represents the absence of the feature in the original input $\mathbf{x}$.

For Integrated Gradients, 64 steps for approximating the integral were used together with a black baseline image.

\textbf{epsilon-LRP.}
Layer-wise relevance propagation (LRP) \cite{Bach.2015} computes an attribution map $R_{LRP} \in \mathbb{R}^{W \times H \times C}$ by propagating the prediction back through the network with the help of specific propagation rules. The so-called epsilon rule is defined as $R_j = \sum_k \frac{a_j w_{jk}}{\sum_j a_j w_{jk} + \epsilon \sign \sum_j a_j w_{jk}}R_k$, where $j$ and $k$ are neurons of two consecutive layers of the neural network and $a_j w_{jk}$ is the weighted activation between the neurons of two consecutive layers.

In this work, $\epsilon$-LRP is parameterized with an $\epsilon$-value of $10$, using the software library \emph{DeepExplain}~\cite{Ancona.2020}.

\subsection{Metrics}\label{metrics}
This Section briefly introduces the three metrics used for the quantitative evaluation of attribution maps' ability to detect bias.

\textbf{Relevance Mass Accuracy.} 
The relevance mass accuracy (RMA), introduced in~\cite{Osman.2020}, is defined as the fraction between the relevance values $R$ that lie inside a ground truth mask $GT$ and all relevance values. Intuitively, RMA measures how much of the relevance's \enquote{mass} lies within the GT area and is defined as 


\vspace{-2mm}
\begin{equation} \label{eq:rma}
    \mathrm{RMA} = {\displaystyle  \sum^{K}_{
        \substack{k=1 \\
        \mathrm{s.t.} \  p_k\ \in \ GT
    }} {R_{pk}}} \cdot z,
\end{equation}

where $z$ is $ \frac{1}{\sum^{N}_{k=1}{R_{pk}}}~$, $N$ is the total number of image pixels, $K$ is the number of pixels within $GT$ and $R_{pk}$ is the relevance value at pixel $p_k$.

\textbf{Relevance Rank Accuracy.}
The relevance rank accuracy (RRA)~\cite{Osman.2020}, measures the fraction of the high intensity relevances that lie within the GT. For computing RRA, a set of top-K relevance locations $S_{top-K} = \{ s_1, s_2, ..., s_K | \newline R_{s_1} > R_{s_2} > .... > R_{s_K} \}$ is obtained, where $K$ is the GT mask's size. Each location $s_p$ is a 2-D vector encoding the horizontal and vertical positions in a pixel grid. Thus, locations in the image that are most relevant for the classifier's prediction are found at the beginning of the set. RRA is defined as
\begin{equation} \label{eq:rra}
    \mathrm{RRA} = \frac{|S_{top-K} \cap GT|}{|GT|}~.
\end{equation}

\textbf{Area over the perturbation curve (AOPC).}
The metric proposed by Samek et al.~\cite{Samek.2017} relies on systematic pixel perturbation, i.e., in several successive steps, the pixel values of the most relevant image regions are perturbed and the effect on the prediction accuracy is observed. This process is quantified as the AOPC and formalized as
\vspace{-4mm}
\begin{equation}
    \aopc = \frac{1}{1+P} \sum_{p=0}^P f(\mathbf{x}'^{(0)}) - f(\mathbf{x}'^{(p)})~,
    \vspace{-2mm}
\end{equation}
where $P$ defines the number of perturbation steps and $\mathbf{x}'^{(p)}$ is the perturbed image after the $p^{th}$ perturbation step. A large AOPC value means that perturbation results in a steep decrease in prediction accuracy, indicating that the attribution method efficiently detects the relevant image regions. In our experiments we follow the approach described in~\cite{Samek.2017} and perform 100 perturbation steps, where we replace the pixel values around a $9 \times 9$ region with random values sampled from a uniform distribution.


\section{Experimental Results}
This section first describes the training process and performance results for the networks used in the experiments. Afterwards, the results for the quantitative bias evaluation are presented.

\subsection{Models}

For the experiments two different CNN architectures are used: EfficientNet-B0~\cite{Tan.2019} and  MobileNetV1~\cite{Howard.17.04.2017}. We train eight different models in total, i.e., one biased and one unbiased network for each architecture and use case, respectively. All networks are trained by fine-tuning, using ImageNet pre-trained models. In order to preserve already trained geometrical features, the first 70 layers of  EfficientNet and the first 26 layers of MobileNet are frozen. For training, a small learning rate ($1e^{-5}$ for \enquote{Cat vs. Dog}, $2e^{-5}$ for  \enquote{Fruits}) and \emph{RMSProp} as optimization algorithm are chosen together with \emph{EarlyStopping} to prevent overfitting. The results are summarized in \Tab{tab:model-results}.

It can be observed that the networks trained on the biased datasets achieve high test accuracies when tested together with the biased datasets, just like the unbiased networks when tested together with the unbiased datasets. Applying the unbiased dataset to the biased networks, though, results in a drop in the prediction accuracy between $17$ and $25$ percent points (\emph{pp}). On the other hand, when applying the biased datasets to the unbiased networks the prediction accuracy is hardly affected---the loss in accuracy ranges between $0$ and $8$ \emph{pp}. Thus, we conclude that the biased networks indeed contain the intended bias, i.e., for \enquote{Cat vs. Dog} the Ball is used as a feature for the Cat class and for \enquote{Fruits} the apple tree background is used as a feature for the Apple class.

\begin{table}[bp]
\setlength{\tabcolsep}{6pt}
\begin{center}
\vspace{-6mm}
\caption{Prediction test accuracy for all networks trained for the two use cases \enquote{Cat vs. Dog} and \enquote{Fruits}.}
\begin{tabular}{l|| c|c | c|c}
     \multirow{2}{*}{Dataset} &\multicolumn{2}{c|}{EfficientNet} & \multicolumn{2}{c}{MobileNet}\\ \cline{2-5}
                                & Bias & No Bias & Bias & No Bias  \\  \hline
      Cat + Ball vs. Dog (Bias) & 0.99 & 0.97    & 0.97 & 0.86     \\
      Cat vs. Dog (No Bias)     & 0.74 & 0.97    & 0.80 & 0.89     \\ \hline
      Fruits (Bias)             & 0.97 & 0.82    & 0.98 & 0.95     \\
      Fruits (No Bias)          & 0.77 & 0.90    & 0.77 & 0.96    
\end{tabular}
\label{tab:model-results}
\end{center}
\end{table}

\subsection{Measure Bias Based on Metrics}
Having confirmed that the CNNs contain the desired bias based on the prediction accuracy, the analysis is continued by means of the attention maps. For this purpose, attribution maps are generated for the different datasets and CNNs by applying the four attribution techniques presented in Section~\ref{sec:attributions}.

Specifically, for each dataset and CNN we first apply the validation (\enquote{Cat vs. Dog}, i.e., 288 images) and test (\enquote{Fruits}, i.e., 810 images) images to the network to get predictions and then generate the attribution maps. We only consider images for which a correct prediction is made. Then, the resulting attribution maps are analyzed by means of the metrics described in Section~\ref{metrics}. For both use cases, the GT areas (Ball, Fruits) are extended by 50\% of their original area.
This is due to the fact that those areas are quite small and some attribution methods provide rather fuzzy attribution maps (SC and GC).

\subsubsection{Cat vs. Dog}
For this use case, we compare the attribution maps created for the four different CNNs based on the \enquote{Cat + Ball vs. Dog} dataset.

\textbf{Relevance Mass and Rank Accuracy.}
\begin{figure}[t]
    \centering
    \includegraphics[width=0.88\textwidth]{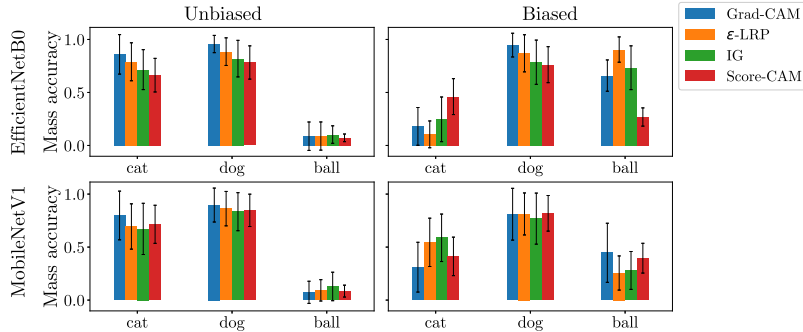}
    \caption{Use Case \enquote{Cat vs. Dog}: Mean RMA and standard deviation (error bars) for \emph{Cat}, \emph{Dog} and \emph{Ball} for all CNNs and attribution techniques.}
    \label{fig:mass-accuracy-imgnet}
\end{figure}
\begin{figure}[t]
    \centering
    \includegraphics[width=0.88\textwidth]{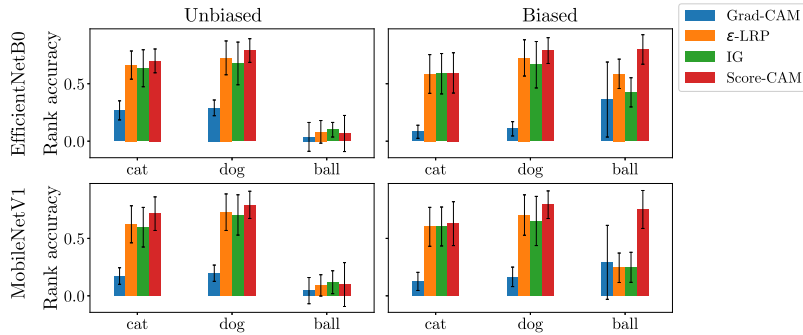}
    \caption{Use Case \enquote{Cat vs. Dog}: Mean RRA and standard deviation (error bars) for \emph{Cat}, \emph{Dog} and \emph{Ball} for all CNNs and attribution techniques.}
    \label{fig:rank-accuracy-imgnet}
    \vspace{-4mm}
\end{figure}
Using the datasets' bounding boxes, RMA and RRA can be computed for the three GT areas \emph{Cat}, \emph{Dog} and \emph{Ball}. The RMA results are displayed in \Fig{fig:mass-accuracy-imgnet}. It is evident that across all attribution techniques the RMA values for \emph{Cat} and \emph{Dog} are very high for the unbiased CNNs, whereas values for \emph{Ball} are close to zero. For the biased networks the RMA values for \emph{Dog} are similar to those of the unbiased networks, while the RMA values for \emph{Cat} have dropped significantly: from $0.7-0.9$ to $0.2-0.5$. At the same time, the values for the \emph{Ball} have increased from approximately $0.05-0.1$ to up to $0.3-0.8$. This indicates that the attention actually shifts from \emph{Cat} to \emph{Ball} for the biased networks, but remains the same for \emph{Dog}. However, this effect is more significant for EfficientNet than for the MobileNet architecture. If those results together with the stronger accuracy drop for the biased EfficientNet (see \Tab{tab:model-results}) are considered, it can be concluded that MobileNet is less biased.
Similar patterns can also be found for RRA (\Fig{fig:rank-accuracy-imgnet}). As for the RMA values, for the biased CNNs the values for \emph{Ball} increase compared to the unbiased CNNs, while the RRA values for \emph{Cat} decrease slightly. A possible explanation could be that the bounding box for \emph{Cat} often covers a large portion of the image. Hence, the chance of relevance values randomly lying within this large image portion is accordingly high. Nevertheless, it can be observed that in the biased networks the attention shifts from the \emph{Cat} to the \emph{Ball}.

Looking at the RMA and RRA values of the attribution methods, an interesting phenomenon can be observed. The RMA values of the different methods are relatively evenly distributed: in most cases the difference is about $0.1$ to $0.2$ points---with GC often achieving the highest values. When considering the RRA results, however, GC underperforms consistently. This indicates that most of its relevance values seem to be within the bounding boxes, but the top-K features are located outside and away from the center of mass.

\begin{table}[t]
\begin{center}
\caption{Use Case \enquote{Cat vs. Dog}: \emph{t}-test between the RMA and RRA values of the biased and unbiased CNNs. The p-value for each GT object and attribution method is shown.}
\begin{tabular}{l | l || cccc | cccc | cccc }
     \multirow{2}{*}{CNN Architecture}             & \multirow{2}{*}{Metric}  & \multicolumn{4}{c|}{Cat}  & \multicolumn{4}{c|}{Dog}   & \multicolumn{4}{c}{Ball}  \\ 
                                  &         & GC & LRP & IG & SC & GC & LRP & IG & SC & GC & LRP & IG & SC \\ \hline
    \multirow{2}{*}{EfficientNet} & RMA     & 0.00 & 0.00 & 0.00 & 0.00 & 0.30 & 0.33 & 0.07 & 0.17 & 0.00 & 0.00 & 0.00 & 0.00 \\
                                  & RRA     & 0.00 & 0.00 & 0.06 & 0.00 & 0.00 & 0.89 & 0.48 & 0.83 & 0.00 & 0.00 & 0.00 & 0.00 \\ \hline
    \multirow{2}{*}{MobileNet}    & RMA     & 0.00 & 0.00 & 0.01 & 0.00 & 0.00 & 0.07 & 0.02 & 0.33 & 0.00 & 0.00 & 0.00 & 0.00 \\
                                  & RRA     & 0.00 & 0.29 & 0.91 & 0.00 & 0.00 & 0.47 & 0.04 & 0.61 & 0.00 & 0.00 & 0.00 & 0.00 \\ 
\end{tabular}
\label{tab:pvalues-imgnet}
\vspace{-6mm}
\end{center}
\end{table}

To determine whether the shift in attention observed using RMA and RRA between the biased and unbiased networks is statistically significant, Welch's \emph{t}-test is performed. For this purpose, the null hypothesis is made that there is no difference between the expected RMA/RRA values of the biased and unbiased CNNs. For both metrics, the test is performed separately for each combination of GT object and attribution method. The results are listed in \Tab{tab:pvalues-imgnet}. We choose the usual significance level $\alpha = 0.05$, which means that if $p \leq \alpha $ the null hypothesis is rejected. It is noticeable that for \emph{Ball} the p-values are always zero, leading to a rejection of the null hypothesis. Except for the RRA values for LRP and IG, this also applies to \emph{Cat}. For \emph{Dog} $p > 0.05$ holds for most cases.

\textbf{AOPC.}
The AOPC results are summarized in \Tab{tab:aopc-catdog}. In order to analyze the effect of the bias, the results are split between the Cat and Dog class. In general, the results show that IG and LRP outperform the other two methods as they consistently achieve higher AOPC values. However, for all methods an unexpected effect can be observed.
For the biased CNNs, IG and LRP get high AOPC values for the Cat class, indicating that the attribution maps efficiently highlight the relevant image regions. Considering the RRA results (high values for Cat and Ball), we can presume the Cat and Ball features are relevant for decision making. In contrast, the AOPC values for the Dog class are relatively small, between $0$ and $0.15$. From the RRA analysis, it is evident that a majority of the relevant pixels for the Dog class lie within the Dog region. However, since perturbing these pixels has no appreciable effect on the prediction accuracy, we can assume that the marked pixels have low relevance for the class decision.
One explanation could be that the induced bias causes the trained models to base their decisions almost entirely on the prominent Ball feature in the Cat class. In comparison, the Dog class' features seem to be irrelevant such that the respective AOPC values do not exceed the random values. This effect seems lifted and even reversed for models trained without bias, turning the Dog class into the dominant one.

\begin{table}[t]
\begin{center}
\caption{Use Case \enquote{Cat vs. Dog}: AOPC values for the systematic perturbation using the attribution maps and a random perturbation as a baseline.}
\begin{tabular}{l|| cc | cc | cc | cc | c}
     \multirow{2}{*}{Network}
     & \multicolumn{2}{c|}{Grad-CAM} & \multicolumn{2}{c|}{$\epsilon$-LRP} & \multicolumn{2}{c|}{IG}  & \multicolumn{2}{c|}{Score-CAM} & \multirow{2}{*}{Random} \\ \cline{2-9}
                   & Cat & Dog & Cat & Dog & Cat & Dog & Cat & Dog &  \\  \hline \hline
      EfficientNet (Bias) & 0.12  & 0.01 & 0.67  & 0.00 & \textbf{0.72}  & 0.00 & 0.11  & 0.01 & 0.04 \\
      MobileNet (Bias) & 0.07  & 0.14 & 0.29  & 0.14 & \textbf{0.35}  & 0.15 & 0.07  & 0.13 & 0.10 \\ \hline
       EfficientNet (No Bias)& 0.02  & 0.35 & 0.05  & 0.20 & 0.04  & 0.18 & 0.00  & \textbf{0.36} & 0.04 \\
       MobileNet (No Bias) & 0.03  & 0.39 & 0.02  & \textbf{0.50} & 0.02  & \textbf{0.50} & 0.02  & 0.34 & 0.17 \\
     \hline
\end{tabular}
\label{tab:aopc-catdog}
\vspace{-6mm}
\end{center}
\end{table}

\begin{figure}[t]
    \centering
    \includegraphics[width=0.9\textwidth]{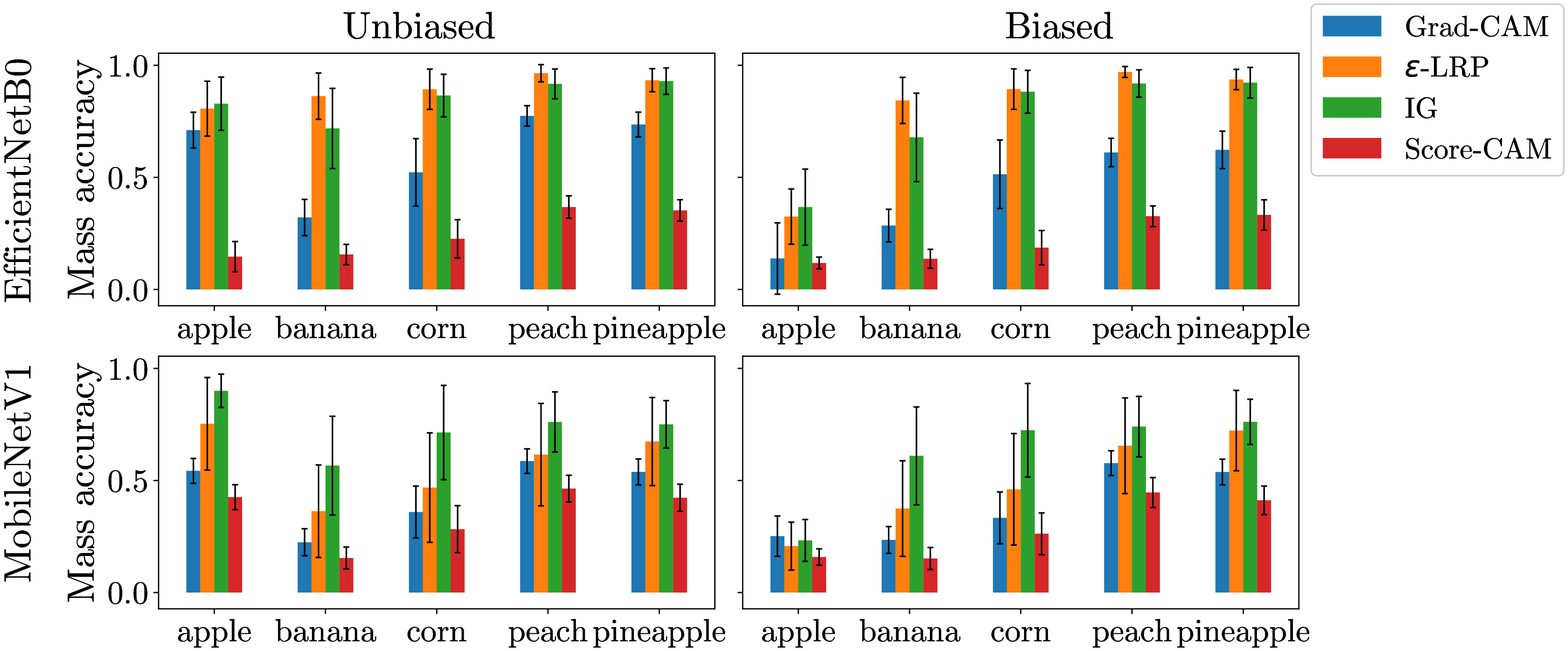}
        \caption{Use Case \enquote{Fruits}: Mean RMA and standard deviation (error bars) for each class for all CNNs and attribution techniques.}
    \label{fig:mass-accuracy-background}
    \vspace{-4mm}
\end{figure}

\begin{figure}[t]
    \centering
    \includegraphics[width=0.89\textwidth]{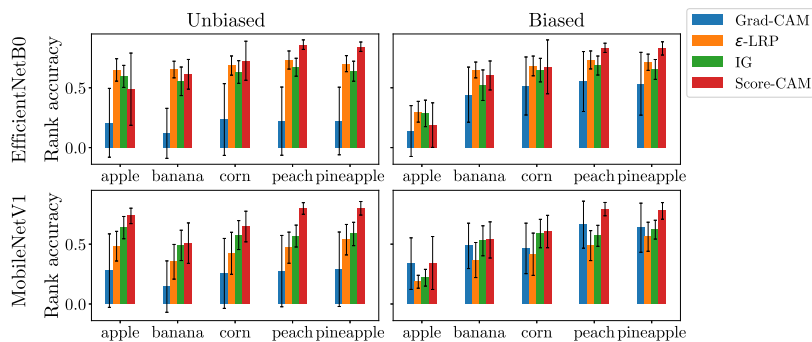}
    \caption{Use Case \enquote{Fruits}: Mean RRA and standard deviation (error bars) for each class for all CNNs and attribution techniques.}
    \label{fig:rank-accuracy-background}
    \vspace{-4mm}
\end{figure}

\subsubsection{Fruits}
\vspace{-2mm}
As for the previous use case, the results for the three metrics RMA, RRA and AOPC are presented.

\textbf{Relevance Mass and Rank Accuracy.}
The RMA and RRA results are summarized in \Fig{fig:mass-accuracy-background} and~\ref{fig:rank-accuracy-background}. Specifically, for EfficientNet an almost equal pronouncement of RMA and RRA values across all five classes can be observed for the unbiased networks. In contrast, for the biased networks, the RMA and RRA values of the Apple class are significantly lower: for RMA the decrease is $\sim50$ \emph{pp}, for RRA $\sim30$ \emph{pp}. For MobileNet, the RMA and RRA vary more significantly between the different classes. Nevertheless, it can also be observed here that the values for Apple are significantly lower for the biased than for the unbiased networks. The results suggest that all attribution maps detect the induced bias diverting the CNNs' attention away from the actual Apple objects towards the diffuse background.

An analysis of the RMA and RRA results provides three interesting observations regarding the performance of the four feature attribution methods. First, as in the \enquote{Cat vs. Dog} use case GC achieves high RMA but comparatively poor RRA values. Second, the exact opposite effect can be observed for SC. For both methods this means that the highest relevance values do not necessarily lie within the mass of relevance values. This effect should also be kept in mind for the visual analysis as those methods usually provide heatmaps that are concentrated strongly on one area. Thus, it is difficult or even impossible to find single outlying pixels with high relevance by visual inspection only. Third, it is noticeable that IG and LRP provide the most stable results. For these methods, usually the highest RMA and RRA values are obtained and a positive correlation between the RMA and RRA values is observable.

\begin{table}[t]
    \centering
    \caption{Use Case \enquote{Fruits}: \emph{t}-test between the RMA and RRA values of the biased and unbiased CNNs. The p-value for each GT object and attribution method is shown.}
    \begin{tabular}{ll||cc | cc | cc | cc | cc}
         CNN & Attribution & \multicolumn{2}{c|}{Apple} & \multicolumn{2}{c|}{Banana} & \multicolumn{2}{c|}{Corn} & \multicolumn{2}{c|}{Peach} & \multicolumn{2}{c}{Pineapple} \\
        Architecture &    Method & RMA & RRA & RMA & RRA & RMA & RRA & RMA & RRA & RMA & RRA  \\ \hline
          \multirow{4}{*}{EfficientNet}  & Grad-CAM & 0.00 & 0.07 & 0.00 & 0.00 & 0.48 & 0.00 & 0.00 & 0.00 & 0.00 & 0.00\\
          & $\epsilon$-LRP & 0.00 & 0.00 & 0.12 & 0.71 & 0.95 & 0.90 & 0.12 & 0.93 & 0.52 & 0.14 \\
          & IG & 0.00 & 0.00 & 0.06 & 0.02 & 0.25 & 0.25 & 0.82 & 0.98 & 0.34 & 0.10 \\
          & Score-CAM & 0.00 & 0.00 & 0.00 & 0.50 & 0.00 & 0.16 & 0.00 & 0.00 & 0.00 & 0.00\\ \hline
          \multirow{4}{*}{MobileNet} & Grad-CAM & 0.00 & 0.04 & 0.13 & 0.00 & 0.30 & 0.00 & 0.13 & 0.00 & 0.94 & 0.00\\
          & $\epsilon$-LRP & 0.00 & 0.00 & 0.66 & 0.41 & 0.46 & 0.44 & 0.11 & 0.21 & 0.02 & 0.07\\
          & IG & 0.00 & 0.00 & 0.09 & 0.01 & 0.90 & 0.45 & 0.16 & 0.84 & 0.33 & 0.00 \\
          & Score-CAM & 0.00 & 0.00 & 0.56 & 0.14 & 0.03 & 0.01 & 0.01 & 0.34 & 0.09 & 0.00 
    \end{tabular}
    \label{tab:pvalues-fruits}
    \vspace{-5mm}
\end{table}

As for the previous use case, we apply a \emph{t}-test to evaluate the statistical significance of the attribution shifts, summarized in \Tab{tab:pvalues-fruits}. Again, it is noticeable that the null hypothesis can be rejected for the biased object, i.e., Apple, since the p-values for each attribution map are zero, except for a single case (GC $+$ RRA).
For EfficientNet, the null hypothesis is mostly rejected for GC and SC, whereas it holds for IG and LRP. The results for MobileNet are more unstable, leading to a rejection of the null hypothesis for several combinations of GT object and attribution method. However, the p-values reflect the results observable in \Fig{fig:mass-accuracy-background} and~\ref{fig:rank-accuracy-background}. 
One explanation for the \enquote{volatile} p-values could be that the bias does not only influence the biased class (Apple) but also other classes which leads to different attribution maps across the biased/non-biased networks. Nevertheless, no other class except Apple shows such a clear shift between the RMA/RRA values of the biased and unbiased networks.

\textbf{AOPC.}
\begin{table}[t]
\begin{center}
\caption{Use Case \enquote{Fruits}: AOPC values for the systematic perturbation using the attribution maps and a random perturbation as a baseline.}
\begin{tabular}{l|| cc | cc | cc | cc | c}
     \multirow{2}{*}{Network}
     & \multicolumn{2}{c|}{Grad-CAM} & \multicolumn{2}{c|}{$\epsilon$-LRP} & \multicolumn{2}{c|}{IG}  & \multicolumn{2}{c|}{Score-CAM} & \multirow{2}{*}{Random} \\ \cline{2-9}
                   & Apple & Rest & Apple & Rest & Apple & Rest & Apple & Rest &  \\  \hline \hline
      EfficientNet (Bias) & 0.03  & \textbf{0.51} & 0.03  & \textbf{0.51} & 0.01  & 0.47 & 0.07  & 0.50 & 0.13 \\
      MobileNet (Bias) & 0.02  & 0.50 & 0.06  & 0.54 & 0.04  & \textbf{0.59} & 0.01  & 0.51 & 0.13 \\ \hline
       EfficientNet (No Bias) & 0.61  & 0.52 & \textbf{0.64}  & 0.54 & 0.63  & 0.50 & 0.49  & 0.53 & 0.03 \\
       MobileNet (No Bias) & 0.75  & 0.50 & 0.83  & 0.54 & \textbf{0.85}  & 0.60 & 0.78  & 0.53 & 0.19 \\
     \hline
\end{tabular}
\label{tab:aopc-fruits}
\vspace{-6mm}
\end{center}
\end{table}
\Tab{tab:aopc-fruits} displays the AOPC results---split between the Apple class and the remaining four classes, named as \enquote{Rest}. For the biased CNNs, perturbation has a nearly no effect on the prediction accuracy for the Apple class, resulting in low AOPC values. More specifically, the systematic perturbation performs worse than the random perturbation. This indicates that the networks do not decide based on small-scale object features, but on large-scale features---here, the apple tree background. For the other classes, however, the AOPC values are much higher compared to random perturbation, suggesting that in this case small image regions are used for decision making, i.e., the fruits themselves. As expected, the attribution maps for the unbiased networks exhibit high AOPC values across all classes, including Apples. This leads to the conclusion that all attribution maps successfully detect the diffuse background bias.


\section{Discussion and Conclusion}

Through the analytical approach followed in this work, we were partly able to measure attribution maps' explanation capability and found quantitative evidence that these techniques detect data biases, both diffuse and focused. Beyond that, the experiments revealed significant differences between the feature attribution methods.
However, the analyses have also shown that attribution maps can sometimes provide misleading explanations.

Specifically, the results obtained for the use case \enquote{Cat vs. Dog} underline some inconsistency between the metrics RMA/RRA and AOPC. From the AOPC values it can be seen that, contrary to human expectations, the CNNs do not always use the features of both classes equally for the binary classification. Instead, they mainly rely on the features of one class.
Consequently, the features of the other class are irrelevant for the classification decision. Nevertheless, even for the \enquote{irrelevant} class, the attribution maps provide results that seem valid at first sight (high RMA and RRA values), but are not confirmed by AOPC results. For this reason, we recommend that the evaluation of attribution maps should not be based on visual inspection and/or one metric alone, especially when applying attribution-map-based methods for industrial use cases. Rather, several different metrics should be used for evaluation for improving the robustness of the results.

The results also motivate a further expansion of the study's scope in three directions: First, new datasets as well as further CNN architectures can be added modularly to generate a scalable and generic test bench for attribution techniques. Second, further investigation into the inconsistency between the RMA/RRA and AOPC metrics should be conducted to investigate why, although the attribution maps highlight certain image regions as relevant, it may be that these are not actually involved for the prediction. Finally, the observation that some CNN architectures might be more vulnerable to bias than others can be studied in detail in future work.

%
\bibliographystyle{splncs04}

\end{document}